\documentclass[conference]{IEEEtran}
\IEEEoverridecommandlockouts

\usepackage{cite}
\usepackage{amsmath,amssymb,amsfonts}
\usepackage{algorithmic}
\usepackage{graphicx}
\usepackage{textcomp}
\usepackage{xcolor}
\usepackage{booktabs}
\def\BibTeX{{\rm B\kern-.05em{\sc i\kern-.025em b}\kern-.08em
    T\kern-.1667em\lower.7ex\hbox{E}\kern-.125emX}}
\begin{document}

\title{Lightweight Diffusion-based Framework\\ for Online Imagined Speech Decoding in Aphasia

\thanks{This work was partly supported by Institute of Information \& Communications Technology Planning \& Evaluation (IITP) grant funded by the Korea government (MSIT) (No. RS-2019-II190079, Artificial Intelligence Graduate School Program (Korea University), No. RS-2021-II-212068, Artificial Intelligence Innovation Hub, and No. RS-2024-00336673, AI Technology for Interactive Communication of Language Impaired Individuals.}
}
 
\author{\IEEEauthorblockN{Eunyeong Ko}
\IEEEauthorblockA{\textit{Dept. of Artificial Intelligence} \\
\textit{Korea University}\\
Seoul, Republic of Korea \\
eunyeong\_ko@korea.ac.kr}
\and
\IEEEauthorblockN{Soowon Kim}
\IEEEauthorblockA{\textit{Dept. of Artificial Intelligence} \\
\textit{Korea University}\\
Seoul, Republic of Korea \\
soowon\_kim@korea.ac.kr}
\and
\IEEEauthorblockN{Ha-Na Jo}
\IEEEauthorblockA{\textit{Dept. of Artificial Intelligence} \\
\textit{Korea University}\\
Seoul, Republic of Korea \\
hn\_jo@korea.ac.kr}
\and
}
\maketitle

\begin{abstract}
Individuals with aphasia experience severe difficulty in real-time verbal communication, while most imagined speech decoding approaches remain limited to offline analysis or computationally demanding models. To address this limitation, we propose a two-session experimental framework consisting of an offline data acquisition phase and a subsequent online feedback phase for real-time imagined speech decoding. The paradigm employed a four-class Korean-language task, including three imagined speech targets selected according to the participant’s daily communicative needs and a resting-state condition, and was evaluated in a single individual with chronic anomic aphasia. Within this framework, we introduce a lightweight diffusion-based neural decoding model explicitly optimized for real-time inference, achieved through architectural simplifications such as dimensionality reduction, temporal kernel optimization, group normalization with regularization, and dual early-stopping criteria. In real-time evaluation, the proposed system achieved 65~\% top-1 and 70~\% top-2 accuracy, with ``Water" class reaching 80~\% top-1 and 100~\% top-2 accuracy. These results demonstrate that real-time–optimized diffusion-based architectures, combined with clinically grounded task design, can support feasible online imagined speech decoding for communication-oriented BCI applications in aphasia.
\end{abstract}


\section{INTRODUCTION}
Individuals with aphasia face profound challenges in establishing direct verbal communication, which significantly constrains their capacity to convey thoughts, needs, and emotional states. Brain-computer interfaces (BCIs) offer a novel therapeutic modality to reinstate communicative function in such populations by facilitating direct decoding of neural signals for device control, thereby circumventing impaired motor and linguistic networks consequent to cerebral injury~\cite{anumanchipalli2019speech, b8, b6, moses2021neuroprosthesis}.

BCIs facilitate device operation through the direct decoding of neural signals, thereby enabling individuals with motor paralysis or language disorders to regain communicative and interactive capabilities~\cite{b15, b9, proix2022imagined}. The interpretation of electroencephalogram (EEG) necessitates the application of sophisticated artificial intelligence models capable of extracting meaningful neural correlates from high-dimensional neural signal data~\cite{b5, b14}. Through the utilization of advanced machine learning and deep learning architectures, BCIs can unlock diverse modalities of communication, significantly expanding the functional repertoire available to affected individuals and substantially enhancing their autonomy and social participation.

Despite the considerable promise of BCIs, contemporary research predominantly relies on offline processing paradigms wherein neural signal acquisition, model training, and classification are executed sequentially in a post-hoc fashion. The transition to real-time, closed-loop BCI systems necessitates stringent constraints on both latency and computational complexity~\cite{meng2022implementation, b12, b2}. These competing demands for rapid response kinetics and reduced model dimensionality present significant technical challenges that must be addressed to realize clinically viable, real-time BCI systems capable of supporting practical communication applications~\cite{angrick2021real}.

In this study, we present a two-session experimental framework designed for individuals with aphasia, consisting of an offline data acquisition phase followed by a real-time online feedback phase. To evaluate the feasibility of subject-specific online decoding under clinically realistic constraints, a single participant with chronic anomic aphasia was recruited. The experimental paradigm was tailored to the participant’s daily communicative needs and implemented as a four-class Korean-language task, including three imagined speech targets and a resting-state condition. EEG data were collected during a 30 minutes data acquisition phase, followed by a 5 minutes real-time online feedback phase to assess closed-loop operation. In addition, we introduce an optimized lightweight variant of a diffusion-based neural decoding architecture tailored for real-time inference through systematic architectural modifications, including dimensionality reduction, temporal kernel optimization, group normalization with regularization, and dual early-stopping criteria to reduce computational overhead while preserving decoding accuracy. The proposed framework achieved over 60~\% classification accuracy in real-time imagined speech decoding from a post-stroke aphasia participant, supporting the clinical feasibility of computationally efficient BCI systems for communication restoration.

\begin{figure*}[t]
\centering
\includegraphics[width=\textwidth]{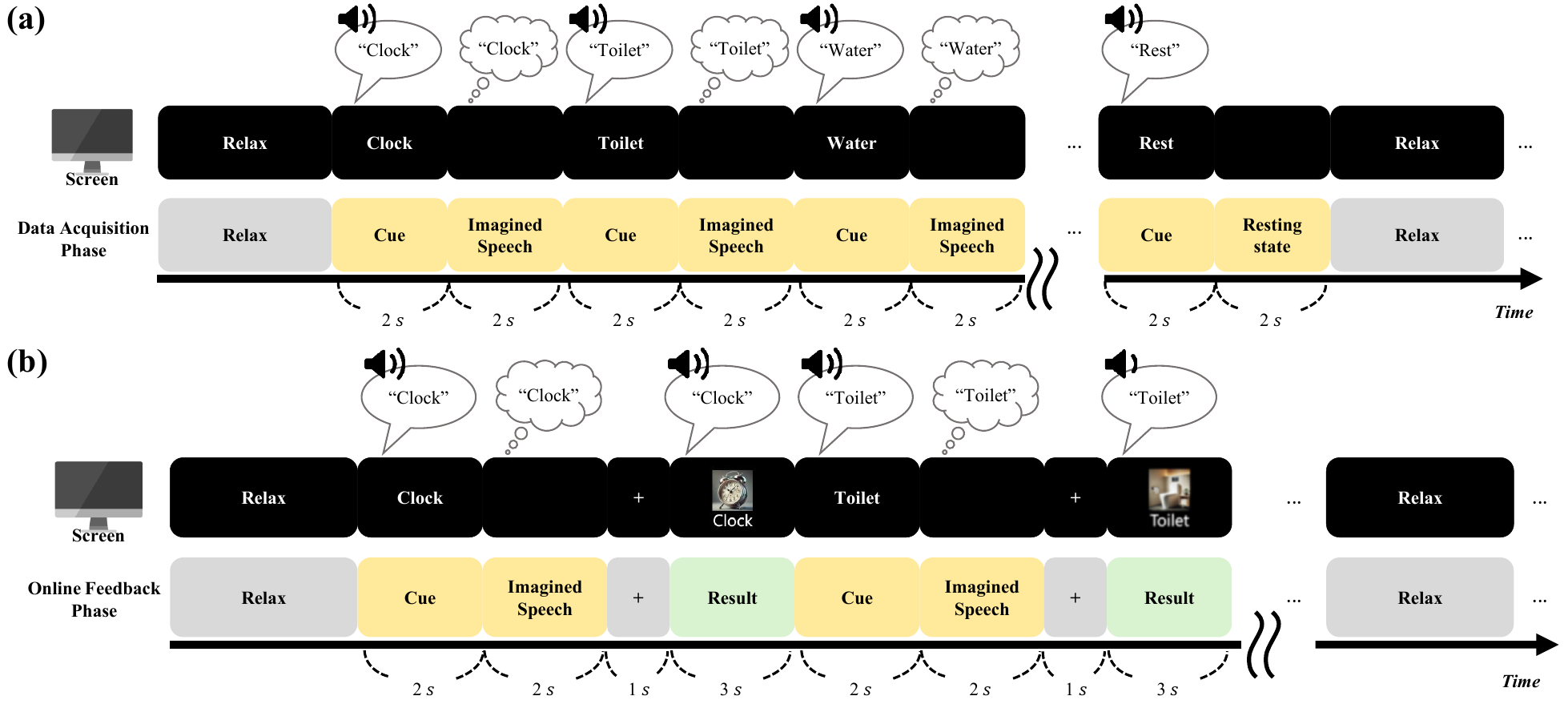}
\caption{Experimental paradigm. (a) Data acquisition phase: 2 seconds cue, imagination, and fixation. (b) Online feedback phase: Identical timing with an added 3 seconds feedback phase using confidence-scaled visual and auditory cues. All sessions occurred in a clinical therapy room with optional caregiver presence.}
\label{fig1}
\end{figure*}

\section{MATERIALS AND METHODS}
\subsection{Experiment Paradigm}
The experimental protocol consisted of two consecutive phases~\cite{b13, b1}. The first was the data acquisition phase, followed by the real-time closed-loop feedback phase. Rest intervals were systematically incorporated to prevent participant fatigue and ensure sustained cognitive engagement. The overall timing and stimulus sequence for each phase are visually depicted in Fig.~\ref{fig1}, with key events and durations.

\subsubsection{Data Acquisition Phase}
During this phase (Fig.~\ref{fig1}a), the participant received simultaneous multimodal cues consisting of textual and auditory stimuli presented for 2 seconds. Upon cue offset, a blank visual display appeared, during which the participant engaged in imagined speech production corresponding to the preceding cue for 2 seconds. This stimulus-imagery cycle was repeated consecutively. The trial structure including relaxation, cue presentation, and imagined speech is sequentially depicted in Fig.~\ref{fig1}a, emphasizing the fixed timing windows and alternating cognitive states. In total, 400 trials were collected, with each class presented 100 times in a fully randomized order across four categories including imagined Korean words ``Clock", ``Toilet", and ``Water", and a resting-state condition, following common experimental practices in EEG-based imagined speech data acquisition~\cite{li2021decoding}. Following completion of 20 successive trials, an extended rest interval was provided prior to resumption of subsequent trial blocks; this rest protocol is also denoted in the Fig.~\ref{fig1}a~\cite{b4}.

\subsubsection{Real-Time Online Feedback Phase}
This phase comprised a sequence of events visually summarized in Fig.~\ref{fig1}b. At the start of each trial, synchronized text and audio cues were presented to the participant for 2 seconds, immediately followed by a blank screen for 2 seconds, during which the participant performed imagined speech production. Subsequently, a fixation cross appeared for approximately 2 seconds, marking the interval in which the participant-specific decoder model executed real-time neural signal classification. Upon completion of decoding, multimodal feedback was delivered for 3 seconds~\cite{hinterberger2004multimodal}. This feedback comprised both visual and auditory presentations that indicated the predicted outcome. In the Fig.~\ref{fig1}b, the gradation of visual contrast and audio intensity represents dynamic scaling based on the decoder's confidence score, illustrating how feedback salience was modulated to reflect prediction certainty. In total, the real-time online feedback phase consisted of 20 trials, which corresponds to a single block under the protocol in which rest intervals are scheduled every 20 trials~\cite{b10}.


\subsection{Participant}
A 54-year-old female, with chronic anomic aphasia and mixed dysarthria following a left hemisphere stroke, participated in this study. Her severe expressive language impairment made her an appropriate candidate for assessing brain–computer interface technology. The study was approved by the Korea University Institutional Review Board [KUIRB-2023-0429-01] and followed the guidelines of the Declaration of Helsinki.

\subsection{Data Acquisition and Preprocessing}
EEG was acquired from 64~scalp electrode sites positioned according to the international 10-20 system using an actiCAP slim electrode cap, sampled at a rate of 500~Hz via the BrainVision Lab Streaming Layer recording interface. Raw EEG data were preprocessed using the Python-based MNE signal processing library~\cite{li2021decoding, b3}. A fifth-order Butterworth infinite impulse response bandpass filter with pass-band frequencies between 0 and 120~Hz was implemented to attenuate non-physiological frequency components and high-frequency noise artifacts~\cite{sun2017fifth}. A complementary notch filter centered at 60~Hz was applied to suppress power line contamination arising from alternating current electrical interference. Baseline normalization was performed by subtracting the mean voltage during the pre-stimulus interval (spanning from -0.2 to 0~seconds relative to imagined speech onset) from the entire epoch. As a final preprocessing step, signals were re-referenced using the common average reference method. This process subtracted the global average voltage from each channel to reduce noise and enhance signal quality~\cite{b7}.

\subsection{Diffusion-Based Neural Decoding Framework}
\begin{figure}[t]
\centering
\includegraphics[width=\columnwidth]{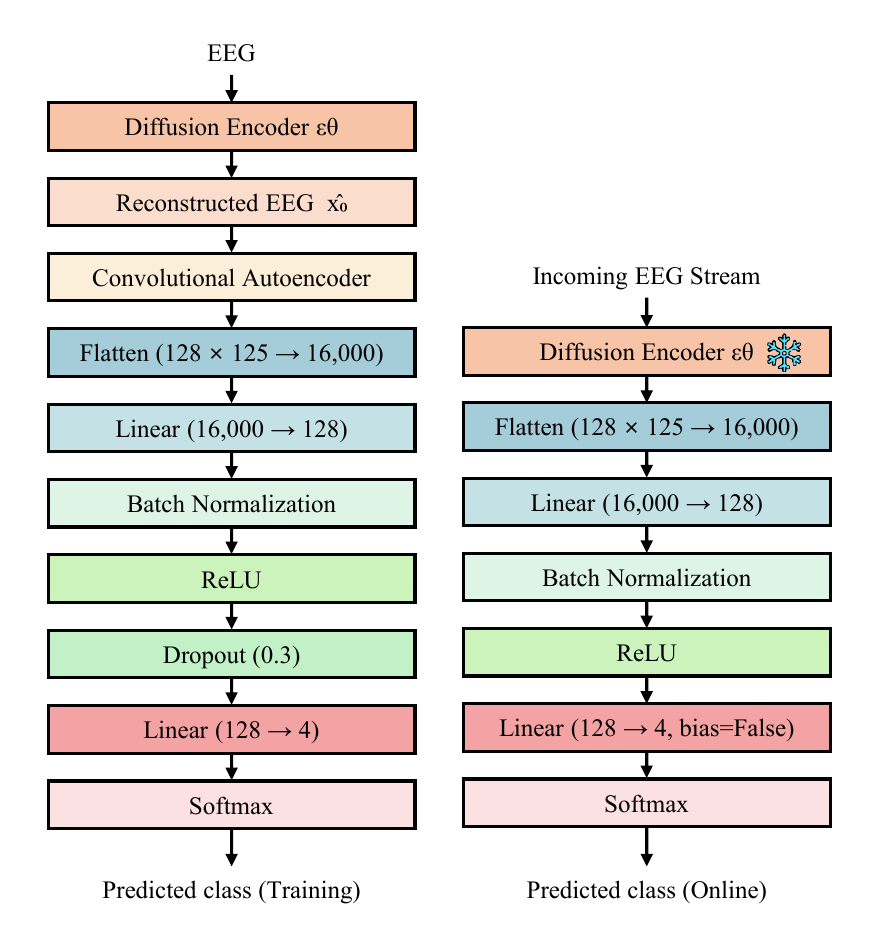}
\caption{Diffusion-based neural decoding framework. (Left) During training, the diffusion encoder and convolutional autoencoder are jointly optimized with dropout regularization. (Right) During real-time decoding, the trained model operates in evaluation mode for low-latency inference.}
\label{fig2}
\end{figure}

\subsubsection{Model Architecture}
We implemented a lightweight diffusion-based framework optimized for rapid training and low-latency inference under limited session data. As illustrated in Fig.~\ref{fig2}, diffusion is formulated as a noise-conditioned representation learning mechanism based on a three-level 1D U-Net backbone~\cite{b11, ronneberger2015u}, enabling robust feature extraction under limited labeled data and session non-stationarity. Time-step and class conditions are incorporated through sinusoidal embeddings and linear class projections, allowing the encoder to adapt feature extraction across noise scales and task contexts. During online operation, the full reverse diffusion sampling process is omitted, and a single denoising-conditioned forward pass with timestep conditioning is performed for low-latency classification using a cosine noise schedule ($s=0.008$, $T=1000$)~\cite{kim2023diff, kim2025eeg}.


\subsubsection{Training Protocol}
Training was performed using Adam with a learning rate of $1 \times 10^{-3}$. Dual-criterion early stopping was adopted to satisfy clinical time constraints. Training was stopped when either the training accuracy exceeded 75~\% or the validation accuracy exceeded 40~\%~\cite{anumanchipalli2019speech, schirrmeister2017deep}. This heuristic criterion was used to accelerate subject-specific calibration while reducing overfitting under limited session data. Per-class confusion matrices were computed to examine systematic class-wise misclassification patterns.

\subsubsection{Real-Time Decoding}
In the online session, the trained diffusion encoder was deployed in a streamlined inference mode to support low-latency operation. Each EEG window was evaluated through a single noise-conditioned forward pass at a fixed inference noise level $\tau_{\mathrm{infer}} = 0.5$, enabling efficient feature extraction for classification without iterative sampling. Continuous EEG was processed in windows of 2~seconds with 0.2~seconds baseline correction, consistent with the offline preprocessing pipeline~\cite{geirnaert2020fast}. The resulting class probabilities were mapped to graded audiovisual feedback by scaling visual clarity and auditory intensity according to prediction confidence~\cite{li2021decoding}. The proposed network contains 14.11~M trainable parameters and achieves an inference time of approximately 2~ms per window of 2~seconds on an NVIDIA Titan XP.

\section{RESULTS AND DISCUSSION}
The proposed framework was assessed with respect to computational efficiency during training and decoding stability during online operation, reflecting the final performance of the experimental protocol.

\begin{table}[t!]
\centering
\tiny
\caption{Performance of real-time imagined speech classification}
\label{tab:performance}
\resizebox{\columnwidth}{!}{%
\begin{tabular}{lcc}
\hline
Class & Top-1 Accuracy & Top-2 Accuracy \\
\hline
Clock & 60~\% & 80~\% \\
Toilet & 20~\% & 80~\% \\
Water & 80~\% & 100~\% \\
Resting state & 40~\% & 60~\% \\
\hline
All & 65~\% & 70~\% \\
\hline
\end{tabular}}
\end{table}

\subsection{Data Acquisition and Model Calibration}
The offline data acquisition phase was conducted over approximately 30 minutes and provided the training data for subject-specific model calibration. Owing to the lightweight design of the diffusion-based encoder and compact classifier, model training and parameter optimization were completed in approximately 20~seconds, enabling rapid deployment for subsequent real-time evaluation.

\subsection{Performance in Real-Time Session Evaluation}
Classification results for the diffusion-based decoding framework under real-time closed-loop conditions are summarized in Tab.~\ref{tab:performance}. The online evaluation, conducted over approximately 5 minutes, achieved 65~\% top-1 and 70~\% top-2 accuracy across four classes. Among the targets, ``Water" showed the most reliable decoding behavior, achieving 80~\% top-1 and 100~\% top-2 accuracy, indicating stable separability and consistent ranking in the real-time setting. In contrast, ``Toilet" exhibited a pronounced ranking discrepancy, suggesting that the correct label was frequently assigned as a near-miss prediction rather than being entirely missed. This pattern is consistent with class-pair ambiguity under real-time non-stationarity, where transient artifacts and variability in imagery consistency can shift decision boundaries within short decoding windows. Importantly, the large top-2 gain for ``Toilet" implies that uncertainty-aware interfaces may remain usable by leveraging multiple ranked candidates rather than relying on a single hard decision.

\subsection{Limitations and Future Directions}
Despite promising closed-loop performance, several limitations remain. Firstly, the pilot cohort limits generalizability, and future work should include additional participants to examine how lesion characteristics and aphasia profiles affect decoding outcomes. Secondly, the current system requires subject-specific calibration, and transfer learning and adaptive fine-tuning may reduce the calibration burden while improving robustness to session variability. Finally, this proof-of-concept evaluated a four-class vocabulary, and expanding the command set while analyzing class-pair confusion patterns will be necessary for practical daily communication support.

\section{CONCLUSIONS}
The objective of this study was to develop a clinically feasible real-time imagined speech decoding framework for individuals with aphasia using a diffusion-based representation and a compact classifier in a two-session design. The results demonstrate that diffusion-trained neural representations can be effectively operationalized for stable online decoding under practical constraints, enabling interactive and confidence-aware feedback in real time. With further refinement, the proposed framework may be applied to real-time command selection interfaces for daily communication support and adaptive neurorehabilitation systems that promote consistent imagined speech through closed-loop interaction.


\end{document}